\documentclass[letterpaper]{article} 
\usepackage{aaai23} 
\usepackage{times}  
\usepackage{helvet}  
\usepackage{courier}  
\usepackage[hyphens]{url}  
\usepackage{graphicx} 
\usepackage{natbib}  
\usepackage{caption} 
\usepackage{algorithm}
\usepackage{algorithmic}

\usepackage{multirow}
\usepackage{listings}
\usepackage{tabularx}
\usepackage{amsmath}
\usepackage{subcaption}
\usepackage{booktabs}
\usepackage{amssymb}
\usepackage{newfloat}
\usepackage{listings}
\DeclareCaptionStyle{ruled}{labelfont=normalfont,labelsep=colon,strut=off} 
\floatstyle{ruled}
\newfloat{listing}{tb}{lst}{}
\floatname{listing}{Listing}
\title{Selective Text Augmentation with Word Roles for Low-Resource \\Text Classification}

\title{Selective Text Augmentation with Word Roles for Low-Resource \\Text Classification}
\author {
        Biyang Guo \thanks{All authors contribute equally and are listed alphabetically.},
        Songqiao Han \footnotemark[1],
        Hailiang Huang\footnotemark[1]\thanks{Corresponding author.}\\
}
\affiliations {
    AI Lab, School of Information Management and Engineering, \\Shanghai University of Finance and Economics,
    Shanghai, China, 200433 \\
    guobiyang2020@gmail.com, \{han.songqiao, hlhuang\}@shufe.edu.cn
}

\usepackage{bibentry}

\begin{document}

\maketitle

\begin{abstract}
  Data augmentation techniques are widely used in text classification tasks to improve the performance of classifiers, especially in low-resource scenarios. Most previous methods conduct text augmentation without considering the different functionalities of the words in the text, which may generate unsatisfactory samples. Different words may play different roles in text classification, which inspires us to strategically select the proper roles for text augmentation. In this work, we first identify the relationships between the words in the text and the text category from the perspectives of \textit{statistical correlation} and \textit{semantic similarity}, and then utilize them to divide the words into four roles --- Gold, Venture, Bonus, and Trivial words, which have different functionalities for text classification. Based on these word roles, we present a new augmentation technique called \textbf{STA} (\textbf{S}elective \textbf{T}ext \textbf{A}ugmentation) where different text-editing operations are selectively applied to words with specific roles.
  STA can generate diverse and relatively clean samples, while preserving the original core semantics, and is also quite simple to implement. Extensive experiments on 5 benchmark low-resource text classification datasets illustrate that augmented samples produced by STA successfully boost the performance of classification models which significantly outperforms previous non-selective methods. Cross-dataset experiments further indicate that STA can help the classifiers generalize better to other datasets than previous methods.
\end{abstract}

\section{Introduction}\label{sec:intro}
Text classification is one of the fundamental tasks in Natural Language Processing (NLP), which has wide applications in news filtering, paper categorization, sentiment analysis, and so on. Plenty of algorithms, especially deep learning models have achieved great success in text classification, such as recurrent neural networks (RNN) \cite{liu2016recurrent, wang2018topic}, convolutional networks (CNN) \cite{kim2014convolutional} and BERT models \cite{devlin2018bert, sanh2019distilbert}. The success of deep learning is usually built on the large training data with good quality, which may not be easily available in real applications. 
Therefore, data augmentation techniques have attracted more and more attention both in academic and industrial communities to improve the generalization ability of text classification models when training data is limited \cite{bayer2021survey}. In this paper, we focus on the low-resource text classification tasks where only small labeled data is available.

\begin{figure}[t]
 \centering
\includegraphics[width=0.45\textwidth]{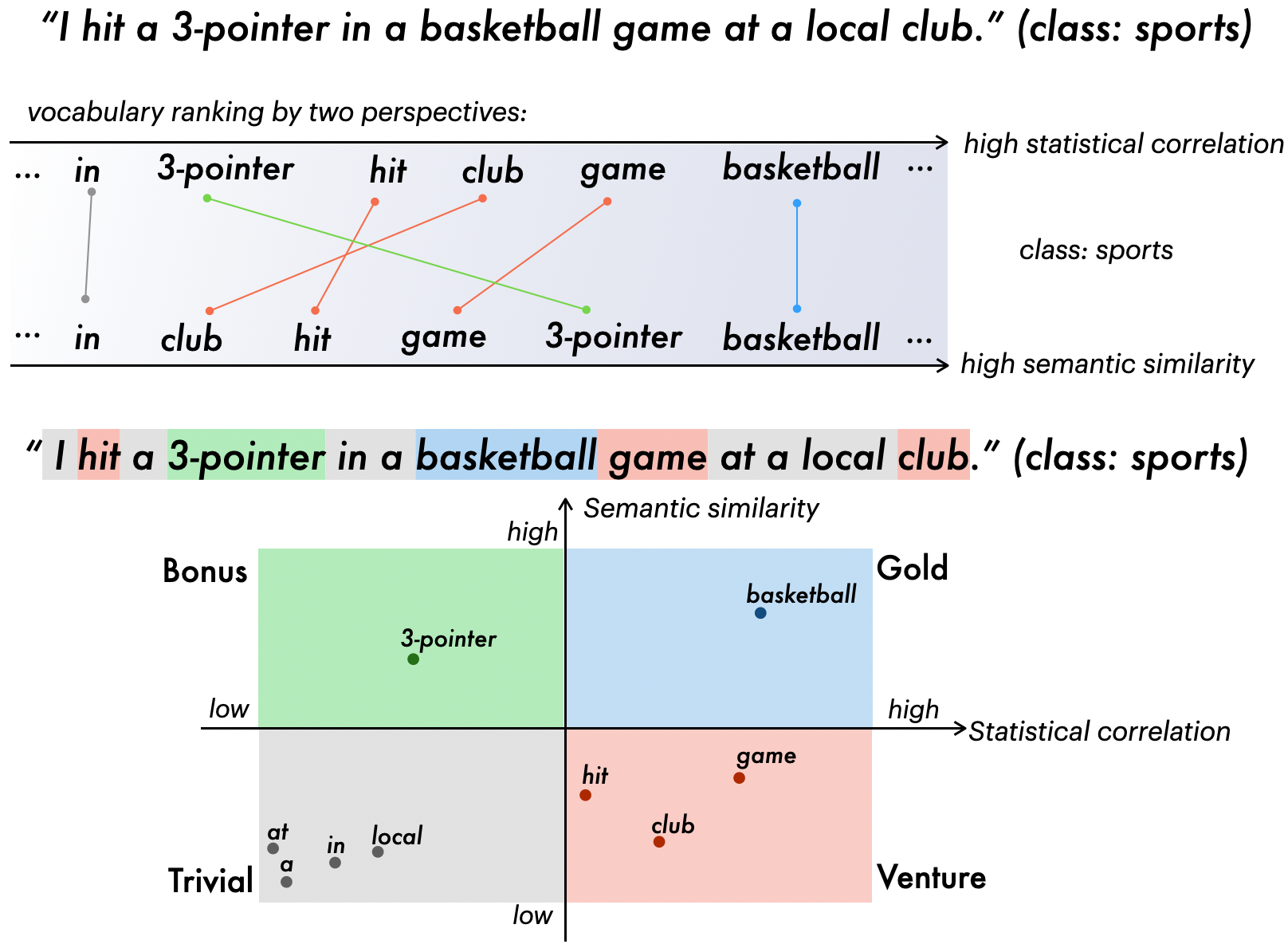} 
 \caption{Illustration of the four word roles}
 \label{roles}
\end{figure}

A lot of text augmentation techniques have been designed to generate more training data. For examples, text-editing methods \cite{wei2019eda, feng-etal-2020-genaug} expand the training data based on simple text-editing rules, such as word replacement and insertion. Back-translation techniques \cite{back_translation, back_translation1} generate new samples by first translating the samples into another language and then back to the original language. Language model based methods\cite{kobayashi2018contextual, lambada, PTM4aug} utilize pretrained language models to synthesize new text for training. However, previous methods mainly augment the text in a \textit{non-selective} manner, without considering the different roles of different words , which may result in undesirable augmented samples: 1) Important class-indicating words may be altered, resulting in some damage to the original meaning or even changing the label of the original text; 2) Unimportant words, noisy words or misleading words may be enhanced after augmentation, which may hurt the generalization ability.



To tackle these issues, we propose to first recognize the \textit{word roles} in the text, and then selectively apply augmentation operations to the words with proper roles. 

Based on our preliminary study, we found that some words with high statistical relevance to a category may have low semantic similarity to that category, and vice versa, as illustrated in Figure \ref{roles}. Motivated by this mismatch, we propose to view the words in text classification tasks from two perspectives. One is the \textit{statistical correlation} which measures how frequent a word co-occurs with certain category while less often with other categories; the other one is the \textit{semantic similarity} which measures the semantic overlap between a word and a category. Intuitively, words with different degrees of statistical correlation and semantic similarity should play different roles in text classification tasks. Therefore, according to the high and low degrees of the two perspectives, the words in the text can be divided into four groups (Figure \ref{roles}) --- Gold, Venture, Bonus, and Trivial words. Based on the recognized word roles, we design \textbf{S}elective \textbf{T}ext \textbf{A}ugmentation (\textbf{STA}) method for text classification tasks which edits the words in the text \textit{selectively} according to their roles during augmentation.



\begin{figure*}[t]
 \centering
\includegraphics[width=\textwidth]{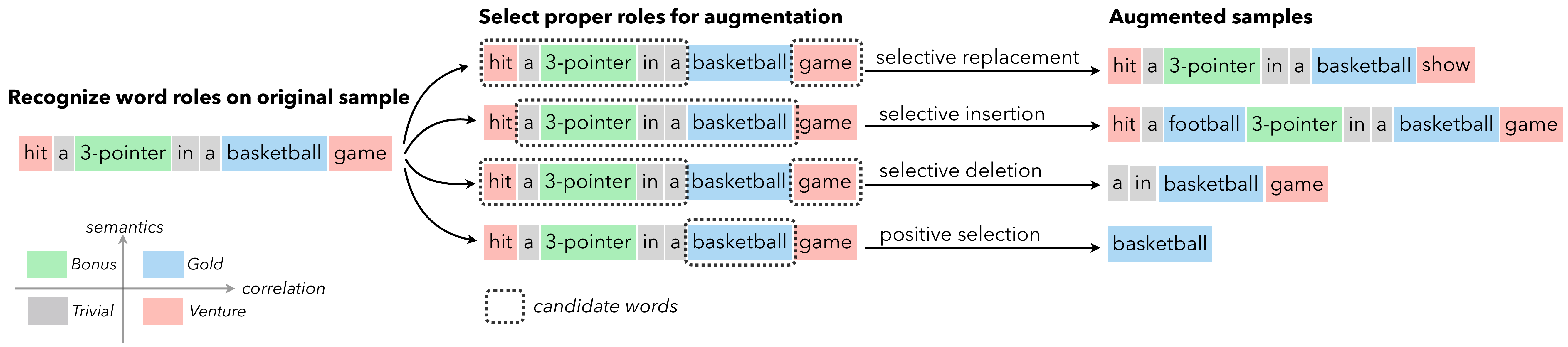} 
\caption{An overview of the process of our proposed selective text augmentation (STA) method. Different text-editing operations are selectively applied to words with specific roles. (Better viewed in color.)}
\label{process}
\end{figure*}

We conduct extensive low-resource text classification experiments on 5 benchmark datasets including 3 topic classification and 2 sentiment analysis datasets, which illustrate that STA significantly outperforms other baselines with an average accuracy improvement at nearly 5\% and 2.6\% with only 50 and 100 training samples respectively. Our proposed STA also surpasses previous augmentation methods by more than 4\% on average in 3 cross-dataset generalization tasks. We conclude our contributions as follows:
\begin{itemize}
    \item[1.] We present the concept of \textbf{word roles} with effective methods to recognize them. The word roles represent the relationships between the words and the categories of the classification tasks considering both \textit{statistical} and \textit{semantic} views, thus are useful for data augmentation and can also inspire related research fields like information retrieval or document representation.
    \item[2.] We propose a new data augmentation approach --- \textbf{STA}, for text classification tasks based on the word roles, which can generate diverse and relatively clean samples while preserving the original core semantics, and is also quite simple to implement and deploy. STA outperforms traditional methods by a large margin on 5 text classification tasks and 3 cross-dataset generalization tasks.
\end{itemize}

\section{Related Work}\label{sec:related-work}

Textual data augmentation techniques are widely used for text classification and other NLP tasks in low-resource scenarios \cite{survey_nlp}. Popupar methods include text editing-based augmentation, such as random word deletion \cite{16-xie2017data, feng-etal-2020-genaug}, synonyms insertion \cite{wei2019eda, feng-etal-2020-genaug}, synonyms replacement (substitution) \cite{sr2011,data-zhang-cnn,sr2015b, feng-etal-2020-genaug} and word order swapping \cite{wei2019eda}; Generative text augmentation, like back-translation \cite{9-xie2019unsupervised,back_translation1,back_translation2} and sentences synthesizing with masked or auto-regressive language models \cite{kobayashi2018contextual,lambada,PTM4aug}; Feature space augmentation \cite{mixup_text,mixup_transformer} which utilize Mixup \cite{zhang2018mixup} for text embeddings during training. Apart from these three types of text augmentation techniques, many other creative methods are proposed such as compositional augmentation \cite{composition,composition2} and adversarial text augmentation \cite{morris2020textattack}.

Recent literature \cite{good_views, da_survey, bayer2021survey, feature_aug} has discussed what properties an ideal text augmentation technique should have. We conclude these properties as follows:
\begin{itemize}
    \item \textit{Diversity}: introduce more diversity to the training set;
    \item \textit{Preservation}: protect the core semantics of the text during augmentation; 
    \item \textit{Filtering}: avoid amplifying the potential noise contained in the original training set. 
    \item \textit{Convenience}: easy to implement and deploy for downstream tasks, even for non-experts.
\end{itemize}

Most previous methods fail to satisfy these properties simultaneously. For example, EDA\cite{wei2019eda} edits the randomly chosen words from the text, which brings more diversity but has risk of damaging the core semantics or even changing the label of the original text. Contextual Augmentation\cite{kobayashi2018contextual} utilizes language model to predict the masked words for substitution based on contextual embeddings, which can generate fluent sentences, but lacks diversity. Back-translation methods\cite{back_translation1,back_translation2} are able to generate diverse and label-preserving samples but usually require large encoder-decoder translation models to guarantee the performance. Generative  methods like \cite{lambada, PTM4aug} can synthesize diverse sentences but require continue pre-training/fine-tuning on down-stream tasks before they can be used. Additionally, none of the above methods try to filter out the potential noise in the text, therefore some misleading words may be enhanced after augmentation, which may hurt the generalization ability. 

In this work, we attempt to design a lightweight text augmentation method, which can generate diverse and clean samples while preserving the core semantics. Therefore, we mainly focus on text-editing-based approach and study how to generate better training samples with simple text-editing operations without the need of pretrained language models.

\section{Methodology}

\subsection{Word Roles Recognition}\label{subsec:roles}

\subsubsection{Word Roles}
Assume that we have a text classification task to train a model to assign one label from $\mathcal{C}=\{c_1,c_2,...,c_k\}$ to a sentence or document. The training set is $\mathcal{D}$ and its vocabulary is $\mathcal{V}$. For a word $w_i \in \mathcal{V}$ and a category $c_j \in \mathcal{C}$, we can examine their relationship from two perspectives:\\
$\bullet$ \textbf{Statistical Correlation}. This measures how frequent the word $w_i$ co-occurs with class $c_j$ while not with other classes in the training set.\\
$\bullet$ \textbf{Semantic Similarity}. This measures how much semantics the word $w_i$ share with the label of class $c_j$.

With the two perspectives, all the words in a training document can be divided into four different roles (Figure \ref{roles}):
\begin{itemize}
    \item[1.] \textbf{Gold} words: these are useful class-indicating words, with \textit{high} statistical correlation and \textit{high} semantic similarity with the corresponding category;
    \item[2.] \textbf{Venture} words: these words usually co-occur frequently with their corresponding categories but have \textit{low} semantic similarities. Thanks to their high statistical correlation with the class, these Venture words can usually bring extra information about the class. However, their low semantic similarity puts them at greater risk to be the noise or misleading words, which are harmful for model training;
    \item[3.] \textbf{Bonus} words: these words have \textit{low} statistical correlation but \textit{high} semantic similarity. Bonus words usually don't occur frequently in the training data, but are quite useful for model's generalization ability;
    \item[4.] \textbf{Trivial} words: these are those with \textit{low} statistical correlation and \textit{low} semantic similarity, which are perhaps less important in model prediction.
\end{itemize}

\subsubsection{Word Roles Recognition}\label{recognition}
To recognize the words of different roles in a dataset, we should decide proper metrics to measure the above two perspectives.

For the measurement of \textit{statistical correlation} with the class, we employ weighted log-likelihood ratio (WLLR) to select the class-correlated words from the text sample. This is inspired by \cite{yu2016learning} where WLLR is used to find out the "pivot words" for sentiment analysis. The WLLR score is computed by:
\begin{align*}
    wllr(w,y) = p(w|y)log(\frac{p(w|y)}{p(w|\bar{y})})
\end{align*}
where $w$ is a word, $y$ is a certain class and $\bar{y}$ represents all the other classes in the classification dataset. $p(w|y)$ and $p(w|\bar{y})$ are the probabilities of observing $w$ in samples labeled with $y$ and with other labels respectively. We use the frequency of a word occurring in the certain class to estimate the probability.

To measure the \textit{semantic similarity} between a word and the meaning of the class label, a straightforward way is to use word vectors pre-trained with skip-gram \cite{w2v-skipgram} or Glove \cite{pennington2014glove}. We are not using transformer-based models like BERT for similarity measuring due to their high inference cost. Some also reveal that static word-embeddings can achieve comparable and even better performance than BERT-like models in similarity measurement tasks, especially in word-level \cite{s-bert}. We compute the cosine similarity between a word and a class label to see their semantic distance:
\begin{align*}
    similarity(w,l) = \frac{v_w \cdot v_l}{\| v_w \|  \| v_l \|}
\end{align*}
where $l$ represents the label and $v_w$, $v_l$ are word vectors for the word $w$ and the label $l$. We can also use a description of the label to obtain $v_l$ by averaging the word vectors of each word in the description for better label representation. In our experiments, we find that simply using the word or phrase of the label itself is enough to measure the similarity between a word and the category.

We compute the WLLR and similarity scores of each word, and set a threshold to divide the \textit{high} and \textit{low} scores (or set different thresholds respectively). We call the words with high (low) WLLR socres as $C_h$ ($C_l$) and words with high (low) similarity socres as $S_h$ ($S_l$). Then we can extract these words with the following rules:
\begin{align*}
    W_{G} &= \{ w| w \in C_h \cap S_h\} \\
    W_{V} &= \{ w| w \in C_h \cap S_l \} \\
    W_{B} &= \{ w| w \in C_l \cap S_h\} \\
    W_{T} &= \{ w| w \in C_l \cap S_l \} 
\end{align*}
where $W_{G}$, $W_{V}$, $W_{B}$ and $W_{T}$ are Gold, Venture, Bonus, and Trivial words respectively. 

Given a training sample, there are two strategies to assign the roles to the words from the text: \textit{global} or \textit{local}. The global strategy determine the thresholds for roles division based on all the words in the vocabulary, while for local strategy the division is based on the words from the current sample. In other words, the local strategy use a relative view to distinguish high and low scores, therefore the a word's role may change when it occurs in another sample.

In our experiments, we use the \textit{median number} as the threshold for the local strategy, and choose the \textit{upper (lower) quartile} as the high (low) score threshold for the global strategy. We find these choices of thresholds to work well in our first try and did not tune these thresholds. Tuning these hyper-parameters might result in further gains for our proposed method. 

\subsubsection{A Real Case for Illustration}
Here we provide an actual example to show the differences of these roles more vividly. We randomly sample 100 documents from the 20 Newsgroup\footnote{https://archive.ics.uci.edu/ml/datasets/Twenty+Newsgroups
} text classification dataset, which consists of 20 categories. We do the word roles recognition using the \textit{global} strategy and some examples are show in Table \ref{tab:example1}. 

\begin{table}[t]
\small
\caption{Real examples of word roles. These words are extracted from the documents of the \textit{electronics} category of the 20 Newsgroup dataset with only 100 samples.}
\begin{tabularx}{\columnwidth}{lX}
\toprule[1.5pt] \multicolumn{2}{l}{category: \textit{electronics}}                                                                            \\
\midrule
\textbf{Gold:}    & {[}"transistor", "Electronics", "circuit", "pulses", "sensor", "batteries", "engineering", ...{]} \\
\textbf{Venture:} & {[}"Caruth", "sufficient", "immediate", "occur", "signature", "firing", "Oliver", "Mr", ...{]}    \\
\textbf{Bonus:}   & {[}"Computer", "systems", "ca", "Engineering", "@", "etc", "world", "system", ...{]}              \\
\textbf{Trivial:} & {[}"had", "were", "On", "was", "article", "Organization", "did", "From", ...{]}        \\
\bottomrule
\end{tabularx}
  \label{tab:example1}
\end{table}

As shown in Table \ref{tab:example1}, Gold words are highly related to their corresponding category, like "transistor", "circuit" and "batteries" for the \textit{electronics} class. Venture words are not semantically related to their classes, but these words co-occur frequently with the corresponding categories in the given dataset. Intuitively, we expect the models to learn class-indicating features from these Gold words, while less from the Venture words, since some of these Venture words might be misleading. For instance, words like "sufficient" and "signature" appear lots of times in \textit{electronics} documents of the dataset, but we shouldn't rely on these words to tell if a document is about electronics or not. However, according to our preliminary experiments, these Venture words are usually learned by the classifiers as evidence for certain categories, and are likely to result in wrong predictions. Bonus words are semantically similar to the categories, but are less commonly found in these categories. Trivial words seem to be uninformative and are less important for prediction. The examples illustrate that our proposed four kinds of word roles are reasonable and can provide insights for text classification and data augmentation. Note that we only use a subset with 100 documents to identify the roles in these examples, more precise roles recognition can be achieved when more data are provided.

\begin{table*}[]
\small
\centering
\caption{Classification accuracy comparison between different text augmentation methods}
  \label{tab:Main-Exp}
\begin{tabular}{lcccccc|cccccc}
\toprule
\multicolumn{1}{l}{}                & \multicolumn{6}{c}{$n = 50$}                                                                             & \multicolumn{6}{c}{$n = 100$}                                                                            \\
\textbf{methods}                    & \textbf{NG}      & \textbf{Yahoo}   & \textbf{BBC}     & \textbf{IMDB}    & \textbf{SST2}    & avg.             & \textbf{NG}      & \textbf{Yahoo}   & \textbf{BBC}     & \textbf{IMDB}    & \textbf{SST2}    & avg.             \\
\midrule
\textit{non-aug}                    & 35.14          & 18.64          & 92.92          & 53.99          & 59.14          & 51.97          & 51.36          & 22.27          & 92.63          & 74.94          & 77.16          & 63.67          \\
\textit{EDA}                & 40.06          & 30.04          & \textbf{93.26}          & 58.43          & 70.32          & 58.42          & 55.49          & 33.71          & 95.42          & 79.88          & 75.12          & 67.92          \\
\textit{EDA-w2v}                    & 42.56          & 29.44          & 93.15          & 60.97          & 72.66          & 59.76          & 56.48          & 33.51          & 95.15          & 79.10          & 76.20          & 68.09          \\
\textit{MLM-Aug}    & 39.18          & 28.59          & 92.49          & 59.92          & 66.00          & 57.24          & 53.86          & 33.89          & 94.22          & 80.44          & 73.70          & 67.22          \\
\textit{Back-Trans}           & 41.96          & 31.31          & 92.74          & 60.30          & 73.56          & 59.98          & 54.85          & \textbf{37.73} & 95.17          & 76.04          & 75.68          & 67.89          \\
\midrule
\textit{\textbf{STA-global (Ours)}} & \textbf{49.41} & \textbf{33.77} & 92.79          & \textbf{71.34} & 76.90          & \textbf{64.84} & \textbf{58.69} & 35.61          & \textbf{95.57} & 81.91          & 77.96          & 69.95          \\
\textit{\textbf{STA-local (Ours)}}  & 46.40          & 32.34          & 93.21 & 68.56          & \textbf{77.86} & 63.68          & 58.57          & 35.75 & 95.48          & \textbf{83.75} & \textbf{79.64} & \textbf{70.64} \\
\bottomrule
\end{tabular}
\end{table*}

\subsection{Selective Text Augmentation}

Based on the recognized word roles, we propose a new text augmentation technique called \textbf{STA} (\textbf{S}elective \textbf{T}ext \textbf{A}ugmentation), which consists of four operations. Given a piece of text from the training set, one of the following text-editing operations will be applied:
\begin{itemize}
    \item \textbf{Selective Replacement}: Select $n$ words, except for the \textit{Gold} words, from the text and then replace these chosen words with their synonyms or similar words. Can be formulated as $candidates=\{w |w \in W_V \cup W_B \cup W_T\}$.
    \item \textbf{Selective Insertion}: Select $n$ words, except for the \textit{Venture} words, from the text and then insert their synonyms into the original text at random place. Can be formulated as $candidates=\{w |w \in W_G \cup W_B \cup W_T\}$.
    \item \textbf{Selective Deletion}: Choose $n$ words, except for the \textit{Gold} words, from the text and then delete them. Can be formulated as $candidates=\{w |w \in W_V \cup W_B \cup W_T\}$.
    \item \textbf{Positive Selection}: Only select the \textit{Gold} words and concatenate them into a new sentence. By doing so, only the most important parts are remained. To make the new sentence more natural, we also randomly insert some Trivial words and punctuation. Can be formulated as $candidates=\{w |w \in W_G \cup W_T \cup punctuation\}$.
\end{itemize}
During augmentation, each operation of STA will be used to generate different kinds of augmented text, as illustrated in Figure \ref{process}. The number $n$ is determined by a hyper-parameter $p$, which represents the augmentation strength. We can also set different strengths to these operations according to the specific tasks.

\textit{Selective Replacement} makes the sample a bit more different while protecting the core semantics; \textit{Selective Insertion} introduces new words into the sentence, while preventing inserting the potential misleading words, by excluding the Venture words from this operation; \textit{Selective Deletion} focuses on the words which have low statistical or semantic relations with the corresponding categories, thus makes the sample cleaner and more category-related; \textit{Positive selection} can be viewed as another type of deletion operation where all the Venture and Bonus words are deleted from the text. By doing so, the potential noise or misleading words are removed, which may help the text classification model to learn the most class-indicating features of the task.

In fact, more sophisticated methods can be designed based on the proposed word roles, such as combining some of the above operations, or utilizing language models to make the generated text more fluent. However, our main purpose in this work is to provide a set of basic, simple and effective augmentation operations, taking into account the different word roles. More creative methods can be further built upon STA to make the augmentation even stronger, which we leave for future work.

\section{Experiments}
\subsection{Setup}\label{subsec:setup}
\textbf{Datasets.\ \ }
We conduct experiments on five widely used benchmark datasets, including two sentiment classification datasets \textbf{SST2} \cite{Socher-sst2}, \textbf{IMDB} \cite{data-imdb} and three topic classification datasets \textbf{Yahoo} Answers \cite{data-zhang-cnn}, \textbf{20NG}\footnote{https://archive.ics.uci.edu/ml/datasets/Twenty+Newsgroups} and \textbf{BBC} news \cite{data-bbc}.

Since we focus on text classification in low-resource scenarios in this work, we randomly sample $n = \{50,100\}$ examples from the training sets of these datasets for the main experiments. We also evaluate our proposed methods with increasing training size ($n = \{50, 100, 500, 1000\}$) for NG and BBC datasets to further illustrate the effectiveness when more labeled data is available. We use the original full test sets of these datasets for evaluation.



\noindent \textbf{Baselines.\ \ }
In our main experiments, we compare the following methods. \textbf{Non-aug}: train the classifier without using any data augmentation techniques;
\textbf{EDA} \cite{wei2019eda}: first \textit{randomly} select some tokens from the original text, and then apply one of the following text-editing operations: synonyms replacement, synonyms insertion, deletion and word position swap;
\textbf{EDA-w2v}: original EDA utilizes WordNet \cite{miller1995wordnet} for synonyms searching, which makes many synonyms inaccessible. Therefore we implemented an alternative using a Word2Vec\cite{w2v-skipgram} model to find similar words;
\textbf{MLM-Aug} \cite{PTM4aug}: utilizing the masked language modeling (MLM) ability of DistilBERT-base \cite{sanh2019distilbert} to replace certain words according to the contextual embedding. We have also tried BERT-base \cite{devlin2018bert} and RoBERTa-base \cite{liu2019roberta} for MLM-Aug but found they are inferior to DistilBERT-base in our experiments;
\textbf{Back-Trans}: translate a sequence into another language and then back into the original language, using pre-trained encoder-decoder translation models. We use the translation models trained on the Tatoeba corpus of four languages (es, zh, ru, de) as the intermediate language \cite{tiedemann-2020-tatoeba-translation-model}.

For our proposed STA, we experiment two strategies of roles assignment: \textbf{STA-global}: using the global strategy for roles assignment; \textbf{STA-local}: using the local strategy for roles assignment.

\noindent \textbf{Settings.\ \ }
In our main experiments, we use DistilBERT-base \cite{sanh2019distilbert} as the backbone for text classifiers, which is a lightweight transformer model distilled from BERT \cite{devlin2018bert}. To verify the effectiveness of our proposed augmentation method on larger models, we also evaluate on BERT-base, discussed in later sections.

We randomly choose 20\% of the training set as the validation set, use the AdamW \cite{adamw} optimizer with learning rate $lr=5e-5$ for training and use early-stopping with patience=10 to choose the best model. We augment the text 4 times for all augmentation methods for fair comparison. We run all experiments with 5 random seeds and report the average performance. We use publicly available Word2Vec model\footnote{https://code.google.com/archive/p/word2vec/} to find the most similar words for synonyms searching in EDA-w2v and STA. For the proportion of words to be changed during augmentation, we experiment three levels of augmentation strength $p=\{5\%,10\%, 20\%\}$ and choose the strength with the highest validation accuracy. If the number of role words are less than $p$ (not enough for selection), we then select random words from the text as supplements. For other baselines, $p$ is set as recommended in original papers.

\subsection{Main Results}
The main results is reported in Table \ref{tab:Main-Exp}. From the results, we can see that all text augmentation methods are beneficial for text classifiers to improve the performance in this low-resource setting. Among all the baselines, EDA-w2v and Back-Trans are competitive methods, which are significantly better than MLM-Aug. The two strategies of our proposed STA both significantly outperform other baselines, with an average accuracy improvement at nearly 5\% over the best baseline (Back-Trans) when n = 50 with STA-global, and 2.6\% over the best baseline (EDA-w2v) when n = 100 with STA-local. In particular, for NG dataset, which is a quite complicated text classification dataset with 20 classes and many sub-classes, STA-global improves the performance by nearly 7\% over EDA-w2v with only 50 initial training examples. For IMDB, a long text sentiment classification task, STA-global achieved a nearly 10\% performance gain over EDA-w2v when n = 50. These results reveal that with the knowledge of word roles and augmenting the text in a selective manner, STA can help the model to achieve better generalization ability than traditional non-selective data augmentation methods. 

Comparing the performances of STA-global and STA-local, we can see that the global strategy performs better when fewer samples are given while the opposite is true for the local strategy. When the data size is too small, assigning the word roles with global strategy can be more accurate, that might be the reason why STA-global achieves relatively better results when train size is only 50.

STA is slightly inferior to EDA for BBC when n = 50, which may result from over-fitting problem, since BBC is a relatively easy task where only 50 training samples can lead to over 90\% test accuracy without any data augmentation. However, with the increase of train size to 100, STA surpasses all other baselines again. Back-Trans performs the best for Yahoo dataset when n = 100 while STA is the second best for this task, but STA beats Back-Trans in all other tasks.

Note that STA is a light-weight text-editing based augmentation method without using any big models. Although word2vec embeddings are needed to compute the semantic similarity during word roles recognition, it is a much faster procedure compared to MLM-Aug and Back-Trans, which use large pre-trained transformer-based models for augmentation.

\begin{table}[h]
\small
\centering
\caption{Experiments with increasing training sizes}
  \label{tab:trian-size}
\begin{tabular}{lcccc}
\toprule
\textbf{NG}      & \multicolumn{1}{l}{$n=50$} & \multicolumn{1}{l}{$n=100$} & \multicolumn{1}{l}{$n=500$} & \multicolumn{1}{l}{$n=1000$} \\
\midrule
\textit{non-aug} & 35.14                  & 51.36                   & 70.88                   & 73.18                    \\
\textit{EDA-w2v} & 42.56                  & 56.48                   & 72.86                   & 74.52                    \\
\textit{\textbf{STA-global}}     & \textbf{49.41}         & \textbf{58.69}          & \textbf{73.37}          & \textbf{75.35}           \\
\bottomrule
\toprule
\textbf{BBC}     & \multicolumn{1}{l}{$n=50$} & \multicolumn{1}{l}{$n=100$} & \multicolumn{1}{l}{$n=500$} & \multicolumn{1}{l}{$n=1000$} \\
\midrule
\textit{non-aug} & \textbf{92.92}         & 92.63                   & 95.07                   & 96.18                    \\
\textit{EDA-w2v} & 93.15                  & 95.15                   & 95.96                   & 97.87                    \\
\textit{\textbf{STA-global}}     & 92.79                  & \textbf{95.57}          & \textbf{96.86}          & \textbf{98.31}          \\
\bottomrule
\end{tabular}
\end{table}

\subsubsection{Experiments with Larger Training Sizes}
We also conduct experiments with increasing training size ($n=\{50, 100, 500, 1000\}$) on the NG and BBC datasets as shown in Table \ref{tab:trian-size}. STA consistently outperforms the competitive EDA-w2v baseline when more labeled data is available.

\begin{figure}[h]
 \centering
\includegraphics[width=0.45\textwidth]{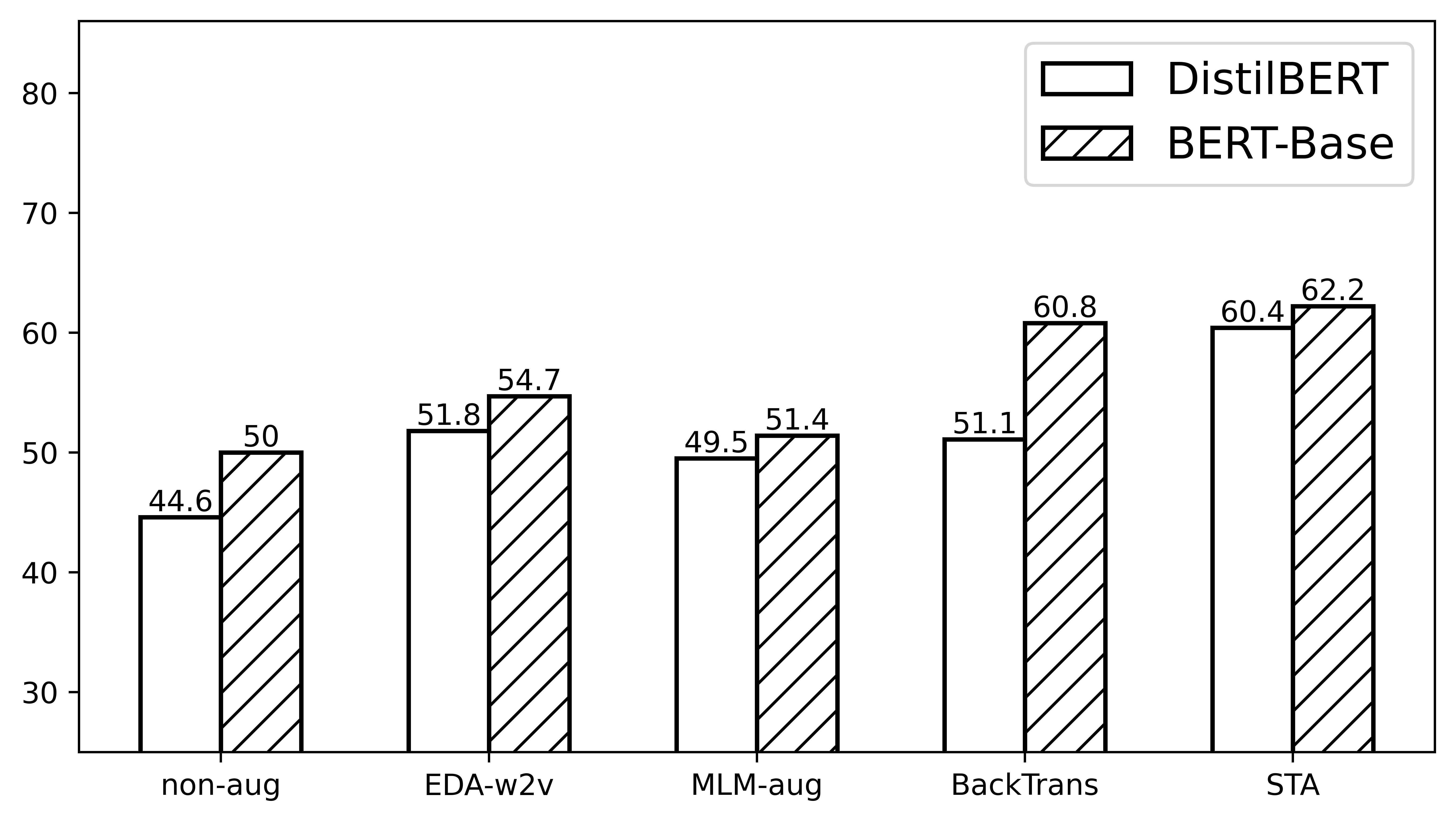} 
 \caption{Evaluation with different classifiers}
 \label{vs}
\end{figure}

\subsubsection{Evaluation with Different Classifiers}\label{sec:different-clf}
To verify the effectiveness of STA on stronger text classification models, we also evaluate different augmentation methods on BERT-base on NG and IMDB datasets and compare their averaged results in Figure \ref{vs}. We can see that STA outperforms other methods both with smaller and larger classification models. Interestingly, even with smaller classification model (DitilBERT), STA helps the model to achieve higher or comparable results than other methods with larger model.

\subsection{Ablation Study}
\subsubsection{Effectiveness of Each STA Operation}
STA is composed of four operations: selective replacement, selective insertion, selective deletion and positive selection. We now evaluate each of these operations, and also compare them with \textit{non-selective} operations where the words are \textit{randomly} selected for text-editing. We conduct experiments on NG, IMDB and Yahoo datasets, and use the \textit{global} strategy for STA. The results are shown in Table \ref{tab:Ablation-1}.

From these experiments, we have two important findings:\\
1. The \textit{positive selection} of STA is particularly powerful as an augmentation operation for low-resource text classification, which is significantly better than all the other operations. The generated new text by positive selection is mainly composed of the Gold words from the text with high statistical and semantic relationships with the label. We speculate that adding these samples into the training set can help the model to learn the most class-indicating features about the class, even if only small number of examples are available.\\
2. \textit{Selective insertion} is more effective than random insertion with very few examples (n = 50) but are less useful when more examples are provided (n = 100). In the contrary, selective deletion and selective replacement are more effective with larger training set. This is probably due to the fact that the word roles recognition is more accurate when providing more labeled data.

These findings can further guide us to design a more powerful selective augmentation method, which we leave for future work.

\begin{table*}[]
\small
\centering
\caption{Effectiveness of each STA operation}
  \label{tab:Ablation-1}
\begin{tabular}{cc|cccc|cccc}
\toprule
\multicolumn{1}{l}{}       & \multicolumn{1}{l}{} & \multicolumn{4}{c}{$n=50$}                                       & \multicolumn{4}{c}{$n=100$}                                      \\
\multicolumn{2}{c|}{\textbf{methods}}                       & \textbf{NG}               & \textbf{IMDB}             & \textbf{Yahoo}            & avg.             & \textbf{NG}               & \textbf{IMDB}             & \textbf{Yahoo}            & avg.             \\
\midrule
\multicolumn{2}{c|}{\textit{non-aug}}                       & 35.14          & 53.99          & 18.64          & 35.92          & 51.36          & 74.94          & 22.27          & 49.52          \\
\midrule
\multirow{3}{*}{\textit{random}}    & \textit{insertion}            & 36.99          & 57.91          & 26.81          & 40.57          & 53.30          & 80.64          & 32.53          & 55.49          \\
                           & \textit{replacement}          & 38.62          & 59.51          & 27.30          & 41.81          & 52.71          & 81.72          & 32.73          & 55.72          \\
                           & \textit{deletion}             & 37.49          & 62.71          & 25.97          & 42.06          & 52.54          & 79.09          & 31.84          & 54.49          \\
\midrule \multirow{4}{*}{\textit{\begin{tabular}[c]{@{}c@{}}selective\\ (Ours)\end{tabular}}} & \textit{insertion}            & 39.13          & 59.95          & 26.83          & 41.97          & 52.87          & 78.21          & \textbf{33.19} & 54.75          \\
                           & \textit{replacement}          & 37.37          & 58.74          & 27.00          & 41.04          & 53.41          & \textbf{82.06}          & 32.78          & 56.09  \\
                           & \textit{deletion}             & 37.12          & 60.65          & 26.56          & 41.44          & 54.41          & 79.65          & 32.66          & 55.57          \\
                           & \textit{positive selection}   & \textbf{40.84} & \textbf{75.09} & \textbf{29.63} & \textbf{48.52} & \textbf{55.32} & 81.73 & 32.88          & \textbf{56.64} \\
                           \bottomrule
\end{tabular}
\end{table*}

\begin{table}[t]
\small
\centering
\caption{Necessity of both statistical \& semantic views for word roles recognition}
  \label{tab:Ablation-2}
\begin{tabular}{lccc}
\toprule
\multicolumn{1}{l}{\textbf{methods}}                   & \textbf{NG}               & \textbf{IMDB}             & avg.             \\
\midrule
\textit{non-aug}                              & 51.36          & 74.94          & 63.15          \\
\textit{EDA}                                  & 55.49          & 79.88          & 67.68          \\
\textit{EDA-w2v}                                  & 56.48          & 79.10          & 67.79          \\
\textit{MLM-Aug}              & 53.86          & 80.44          & 67.15          \\
\textit{Back-Trans}                     & 54.85          & 76.04          & 65.44          \\
\midrule
\textit{STA}                                  & \textbf{58.69} & \textbf{83.75} & \textbf{71.22} \\
\multicolumn{1}{r}{\textit{\textbf{ \ \ \ STA w/o Cor.}}} & 56.53          & 82.66          & 69.59          \\
\multicolumn{1}{r}{\textit{\textbf{ \ \ \ STA w/o Sim.}}} & 55.92          & 80.41          & 68.17         \\
\bottomrule
\end{tabular}
\end{table}


\subsubsection{Necessity of Both Statistical \& Semantic Perspectives for Word Roles Recognition}

Recall that STA is based on word roles recognition, which relies on two different relationship measurements between a word and a class: statistical correlation and semantic similarity. 
It is natural to ask: \textit{do we really need both statistical correlation and semantic similarity with the class for defining word roles?} To verify this, we conduct ablation experiments of by designing two variants of STA:\\
1. \textbf{STA w/o Cor.}: During role words recognition, we only use semantic similarity to separate the words in the text into two types: high-similarity and low-similarity words. By doing so, Gold and Bonus words, or Venture and Trivial words can no longer be distinguished from each other. \\
2. \textbf{STA w/o Sim.}: During role words recognition, we only use statistical correlation to separate the words in the text into two types: high-correlation and low-correlation words. This way, the previous Gold and Venture words are combined together, Trivial and Bonus words are combined together. 

We then follow the same word selection rules with the original STA, and experiment on NG and IMDB datasets with 100 training examples. For NG, STA uses global strategy while for IMDB STA uses local strategy. The results are reported in Table \ref{tab:Ablation-2}. The results illustrate that removing the statistical or semantic factor both lead to performance degradation compared to original STA for NG and IMDB tasks. \textbf{STA w/o Sim.} is worse than \textbf{STA w/o Cor.} which shows that semantic similarity plays a more important part than statistical correlation for proper word selection in STA. Though inferior to the original STA, \textbf{STA w/o Sim.} and \textbf{STA w/o Cor.} both surpass the performance of other non-selective baselines, which illustrates that word selection is vital for text augmentation.

\subsection{Cross-Dataset Generalization Tasks}



To further evaluate the model's generalization ability, we also design 3 groups of cross-dataset generalization tasks, following the experimental design in \cite{DPR} and \cite{PTM-OOD}, where the model is first trained in one dataset and then directly applied to a different dataset without additional fine-tuning. We choose three news classification datasets, \textbf{BBC}, \textbf{FD} \footnote{http://www.nlpir.org/} and \textbf{TH} \footnote{http://thuctc.thunlp.org/} which share two common categories: \textit{politics} and \textit{sports}. Specifically, A$\Rightarrow$B means the model is trained on A dataset and evaluated on the shared classes of B dataset without additional fine-tuning. If B is of a different language, we will first translate B into the same language of A using the open-sourced translation models \cite{tiedemann-2020-tatoeba-translation-model}. We randomly sample 500 examples as the training set from each source dataset and use the test set of the target dataset for evaluation.
The results are reported in Table \ref{cross} which show that STA not only performs well on in-dataset prediction, but also on cross-dataset generalization tasks. We can see that the improvement of STA over EDA-w2v is more than 4\% on average, which shows that the classifiers trained on augmented datasets by STA have learned more common and general features of the shared categories \textit{politics} and \textit{sports} than baselines. These general features are essential for cross-dataset generalization, where the data distribution of the train and test sets may be different. The selective operations of STA like positive selection and selective deletion are especially useful for denoising the dataset while enhancing the core semantics of the categories, thus are helpful for learning the general features of the categories.

\begin{table}[]
\small
\centering
 \caption{Cross-dataset generalization tasks}
\label{cross}
\begin{tabular}{lccc}
\toprule
\textbf{Methods}              & \textit{non-aug}                        & \textit{EDA-w2v}                     & \textbf{\textit{STA}}                         \\
\midrule
\textbf{FD$\Rightarrow$TH}  & 75.62                     & \textbf{76.54}            & 75.31                     \\
\textbf{TH$\Rightarrow$FD}  & 38.50                     & 39.93                     & \textbf{46.68}            \\\midrule[0.5pt]
\textbf{FD$\Rightarrow$BBC} & 74.25                     & 77.24                     & \textbf{82.66}            \\
\textbf{BBC$\Rightarrow$FD} & 34.12                     & 38.38                     & \textbf{41.44}            \\\midrule[0.5pt]
\textbf{TH$\Rightarrow$BBC} & 45.53                     & 46.34                     & \textbf{52.03}            \\
\textbf{BBC$\Rightarrow$TH} & 70.86                     & 69.81                     & \textbf{74.57}            \\
\midrule
\multicolumn{1}{l}{avg.}       & \multicolumn{1}{c}{56.48} & \multicolumn{1}{c}{58.04} & \multicolumn{1}{c}{\textbf{62.12}} \\
\bottomrule
\end{tabular}
\end{table}

\section{Conclusion}
In this work, we first present four types of \textit{word roles} based on two important perspectives to measure the relationships between words and categories. We then propose a new data augmentation technique STA for text classification, which selectively edits the text according to the word roles to preserve the original semantics while introducing more diversity into the training set. Extensive experiments on 5 benchmark classification datasets and 3 groups of cross-dataset generalization tasks illustrate that STA helps the classifier acquire stronger generalization ability over other non-selective augmentation baselines.

Our proposed STA is mainly based on basic text-editing operations, more advanced methods can be designed base on STA, such as utilizing pre-trained language models to make the generated text more fluent. In addition, the idea of word roles can also be applied to other tasks like information retrieval, document representation, and even image classification (where we can study the roles of super-pixels), which we will explore in future work.

\bibliography{aaai23}

\begin{thebibliography}{42}
\providecommand{\natexlab}[1]{#1}

\bibitem[{Anaby-Tavor et~al.(2020)Anaby-Tavor, Carmeli, Goldbraich, Kantor,
  Kour, Shlomov, Tepper, and Zwerdling}]{lambada}
Anaby-Tavor, A.; Carmeli, B.; Goldbraich, E.; Kantor, A.; Kour, G.; Shlomov,
  S.; Tepper, N.; and Zwerdling, N. 2020.
\newblock Do not have enough data? Deep learning to the rescue!
\newblock In \emph{Proceedings of the AAAI Conference on Artificial
  Intelligence}, volume~34, 7383--7390.

\bibitem[{Andreas(2019)}]{composition2}
Andreas, J. 2019.
\newblock Good-enough compositional data augmentation.
\newblock \emph{arXiv preprint arXiv:1904.09545}.

\bibitem[{Bayer, Kaufhold, and Reuter(2021)}]{bayer2021survey}
Bayer, M.; Kaufhold, M.-A.; and Reuter, C. 2021.
\newblock A survey on data augmentation for text classification.
\newblock \emph{arXiv preprint arXiv:2107.03158}.

\bibitem[{Chen et~al.(2021)Chen, Tam, Raffel, Bansal, and Yang}]{survey_nlp}
Chen, J.; Tam, D.; Raffel, C.; Bansal, M.; and Yang, D. 2021.
\newblock An Empirical Survey of Data Augmentation for Limited Data Learning in
  NLP.
\newblock \emph{arXiv preprint arXiv:2106.07499}.

\bibitem[{Devlin et~al.(2019)Devlin, Chang, Lee, and
  Toutanova}]{devlin2018bert}
Devlin, J.; Chang, M.-W.; Lee, K.; and Toutanova, K. 2019.
\newblock BERT: Pre-training of Deep Bidirectional Transformers for Language
  Understanding.
\newblock In \emph{NAACL-HLT (1)}.

\bibitem[{DeVries and Taylor(2017)}]{feature_aug}
DeVries, T.; and Taylor, G.~W. 2017.
\newblock Dataset augmentation in feature space.
\newblock \emph{arXiv preprint arXiv:1702.05538}.

\bibitem[{Feng et~al.(2020)Feng, Gangal, Kang, Mitamura, and
  Hovy}]{feng-etal-2020-genaug}
Feng, S.~Y.; Gangal, V.; Kang, D.; Mitamura, T.; and Hovy, E. 2020.
\newblock {G}en{A}ug: Data Augmentation for Finetuning Text Generators.
\newblock In \emph{Proceedings of Deep Learning Inside Out (DeeLIO): The First
  Workshop on Knowledge Extraction and Integration for Deep Learning
  Architectures}, 29--42. Online: Association for Computational Linguistics.

\bibitem[{Feng et~al.(2021)Feng, Gangal, Wei, Chandar, Vosoughi, Mitamura, and
  Hovy}]{da_survey}
Feng, S.~Y.; Gangal, V.; Wei, J.; Chandar, S.; Vosoughi, S.; Mitamura, T.; and
  Hovy, E. 2021.
\newblock A Survey of Data Augmentation Approaches for NLP.
\newblock In \emph{Findings of the Association for Computational Linguistics:
  ACL-IJCNLP 2021}, 968--988.

\bibitem[{Greene and Cunningham(2006)}]{data-bbc}
Greene, D.; and Cunningham, P. 2006.
\newblock Practical Solutions to the Problem of Diagonal Dominance in Kernel
  Document Clustering.
\newblock In \emph{Proc. 23rd International Conference on Machine learning
  (ICML'06)}, 377--384. ACM Press.

\bibitem[{Guo, Mao, and Zhang(2019)}]{mixup_text}
Guo, H.; Mao, Y.; and Zhang, R. 2019.
\newblock Augmenting data with mixup for sentence classification: An empirical
  study.
\newblock \emph{arXiv preprint arXiv:1905.08941}.

\bibitem[{Hendrycks et~al.(2020)Hendrycks, Liu, Wallace, Dziedzic, Krishnan,
  and Song}]{PTM-OOD}
Hendrycks, D.; Liu, X.; Wallace, E.; Dziedzic, A.; Krishnan, R.; and Song, D.
  2020.
\newblock Pretrained Transformers Improve Out-of-Distribution Robustness.
\newblock In \emph{Proceedings of the 58th Annual Meeting of the Association
  for Computational Linguistics}, 2744--2751.

\bibitem[{Jia and Liang(2016)}]{composition}
Jia, R.; and Liang, P. 2016.
\newblock Data recombination for neural semantic parsing.
\newblock \emph{arXiv preprint arXiv:1606.03622}.

\bibitem[{Karpukhin et~al.(2020)Karpukhin, Oguz, Min, Lewis, Wu, Edunov, Chen,
  and Yih}]{DPR}
Karpukhin, V.; Oguz, B.; Min, S.; Lewis, P.; Wu, L.; Edunov, S.; Chen, D.; and
  Yih, W.-t. 2020.
\newblock Dense Passage Retrieval for Open-Domain Question Answering.
\newblock In \emph{Proceedings of the 2020 Conference on Empirical Methods in
  Natural Language Processing (EMNLP)}, 6769--6781.

\bibitem[{Kim(2014)}]{kim2014convolutional}
Kim, Y. 2014.
\newblock Convolutional Neural Networks for Sentence Classification.
\newblock In \emph{Proceedings of the 2014 Conference on Empirical Methods in
  Natural Language Processing ({EMNLP})}, 1746--1751. Doha, Qatar: Association
  for Computational Linguistics.

\bibitem[{Kobayashi(2018)}]{kobayashi2018contextual}
Kobayashi, S. 2018.
\newblock Contextual augmentation: Data augmentation by words with paradigmatic
  relations.
\newblock \emph{arXiv preprint arXiv:1805.06201}.

\bibitem[{Kolomiyets, Bethard, and Moens(2011)}]{sr2011}
Kolomiyets, O.; Bethard, S.; and Moens, M.-F. 2011.
\newblock Model-portability experiments for textual temporal analysis.
\newblock In \emph{Proceedings of the 49th annual meeting of the association
  for computational linguistics: human language technologies}, volume~2,
  271--276. ACL; East Stroudsburg, PA.

\bibitem[{Kumar, Choudhary, and Cho(2020)}]{PTM4aug}
Kumar, V.; Choudhary, A.; and Cho, E. 2020.
\newblock Data Augmentation using Pre-trained Transformer Models.
\newblock In \emph{Proceedings of the 2nd Workshop on Life-long Learning for
  Spoken Language Systems}, 18--26.

\bibitem[{Liu, Qiu, and Huang(2016)}]{liu2016recurrent}
Liu, P.; Qiu, X.; and Huang, X. 2016.
\newblock Recurrent neural network for text classification with multi-task
  learning.
\newblock \emph{arXiv preprint arXiv:1605.05101}.

\bibitem[{Liu et~al.(2019)Liu, Ott, Goyal, Du, Joshi, Chen, Levy, Lewis,
  Zettlemoyer, and Stoyanov}]{liu2019roberta}
Liu, Y.; Ott, M.; Goyal, N.; Du, J.; Joshi, M.; Chen, D.; Levy, O.; Lewis, M.;
  Zettlemoyer, L.; and Stoyanov, V. 2019.
\newblock Roberta: A robustly optimized bert pretraining approach.
\newblock \emph{arXiv preprint arXiv:1907.11692}.

\bibitem[{Loshchilov and Hutter(2017)}]{adamw}
Loshchilov, I.; and Hutter, F. 2017.
\newblock Decoupled weight decay regularization.
\newblock \emph{arXiv preprint arXiv:1711.05101}.

\bibitem[{Maas et~al.(2011)Maas, Daly, Pham, Huang, Ng, and Potts}]{data-imdb}
Maas, A.~L.; Daly, R.~E.; Pham, P.~T.; Huang, D.; Ng, A.~Y.; and Potts, C.
  2011.
\newblock Learning Word Vectors for Sentiment Analysis.
\newblock In \emph{Proceedings of the 49th Annual Meeting of the Association
  for Computational Linguistics: Human Language Technologies}, 142--150.
  Portland, Oregon, USA: Association for Computational Linguistics.

\bibitem[{Mikolov et~al.(2013)Mikolov, Chen, Corrado, and Dean}]{w2v-skipgram}
Mikolov, T.; Chen, K.; Corrado, G.; and Dean, J. 2013.
\newblock Efficient estimation of word representations in vector space.
\newblock \emph{arXiv preprint arXiv:1301.3781}.

\bibitem[{Miller(1995)}]{miller1995wordnet}
Miller, G.~A. 1995.
\newblock WordNet: a lexical database for English.
\newblock \emph{Communications of the ACM}, 38(11): 39--41.

\bibitem[{Morris et~al.(2020)Morris, Lifland, Yoo, Grigsby, Jin, and
  Qi}]{morris2020textattack}
Morris, J.~X.; Lifland, E.; Yoo, J.~Y.; Grigsby, J.; Jin, D.; and Qi, Y. 2020.
\newblock Textattack: A framework for adversarial attacks, data augmentation,
  and adversarial training in nlp.
\newblock \emph{arXiv preprint arXiv:2005.05909}.

\bibitem[{Pennington, Socher, and Manning(2014)}]{pennington2014glove}
Pennington, J.; Socher, R.; and Manning, C.~D. 2014.
\newblock Glove: Global vectors for word representation.
\newblock In \emph{Proceedings of the 2014 conference on empirical methods in
  natural language processing (EMNLP)}, 1532--1543.

\bibitem[{Reimers and Gurevych(2019)}]{s-bert}
Reimers, N.; and Gurevych, I. 2019.
\newblock Sentence-bert: Sentence embeddings using siamese bert-networks.
\newblock \emph{arXiv preprint arXiv:1908.10084}.

\bibitem[{Sanh et~al.(2019)Sanh, Debut, Chaumond, and
  Wolf}]{sanh2019distilbert}
Sanh, V.; Debut, L.; Chaumond, J.; and Wolf, T. 2019.
\newblock DistilBERT, a distilled version of BERT: smaller, faster, cheaper and
  lighter.
\newblock \emph{arXiv preprint arXiv:1910.01108}.

\bibitem[{Sennrich, Haddow, and Birch(2016)}]{back_translation}
Sennrich, R.; Haddow, B.; and Birch, A. 2016.
\newblock Improving Neural Machine Translation Models with Monolingual Data.
\newblock In \emph{Proceedings of the 54th Annual Meeting of the Association
  for Computational Linguistics (Volume 1: Long Papers)}, 86--96.

\bibitem[{Silfverberg et~al.(2017)Silfverberg, Wiemerslage, Liu, and
  Mao}]{back_translation2}
Silfverberg, M.; Wiemerslage, A.; Liu, L.; and Mao, L.~J. 2017.
\newblock Data augmentation for morphological reinflection.
\newblock In \emph{Proceedings of the CoNLL SIGMORPHON 2017 Shared Task:
  Universal Morphological Reinflection}, 90--99.

\bibitem[{Socher et~al.(2013)Socher, Perelygin, Wu, Chuang, Manning, Ng, and
  Potts}]{Socher-sst2}
Socher, R.; Perelygin, A.; Wu, J.; Chuang, J.; Manning, C.~D.; Ng, A.~Y.; and
  Potts, C. 2013.
\newblock Recursive deep models for semantic compositionality over a sentiment
  treebank.
\newblock In \emph{Proceedings of the 2013 conference on empirical methods in
  natural language processing}, 1631--1642.

\bibitem[{Sun et~al.(2020)Sun, Xia, Yin, Liang, Yu, and He}]{mixup_transformer}
Sun, L.; Xia, C.; Yin, W.; Liang, T.; Yu, P.~S.; and He, L. 2020.
\newblock Mixup-Transformer: Dynamic Data Augmentation for NLP Tasks.
\newblock \emph{arXiv preprint arXiv:2010.02394}.

\bibitem[{Tian et~al.(2020)Tian, Sun, Poole, Krishnan, Schmid, and
  Isola}]{good_views}
Tian, Y.; Sun, C.; Poole, B.; Krishnan, D.; Schmid, C.; and Isola, P. 2020.
\newblock What makes for good views for contrastive learning?
\newblock \emph{Advances in Neural Information Processing Systems}, 33:
  6827--6839.

\bibitem[{Tiedemann(2020)}]{tiedemann-2020-tatoeba-translation-model}
Tiedemann, J. 2020.
\newblock The {T}atoeba {T}ranslation {C}hallenge {--} {R}ealistic Data Sets
  for Low Resource and Multilingual {MT}.
\newblock In \emph{Proceedings of the Fifth Conference on Machine Translation},
  1174--1182. Online: Association for Computational Linguistics.

\bibitem[{Wang et~al.(2018)Wang, Gan, Wang, Shen, Huang, Ping, Satheesh, and
  Carin}]{wang2018topic}
Wang, W.; Gan, Z.; Wang, W.; Shen, D.; Huang, J.; Ping, W.; Satheesh, S.; and
  Carin, L. 2018.
\newblock Topic compositional neural language model.
\newblock In \emph{International Conference on Artificial Intelligence and
  Statistics}, 356--365.

\bibitem[{Wang and Yang(2015)}]{sr2015b}
Wang, W.~Y.; and Yang, D. 2015.
\newblock That’s so annoying!!!: A lexical and frame-semantic embedding based
  data augmentation approach to automatic categorization of annoying behaviors
  using\# petpeeve tweets.
\newblock In \emph{Proceedings of the 2015 Conference on Empirical Methods in
  Natural Language Processing}, 2557--2563.

\bibitem[{Wei and Zou(2019)}]{wei2019eda}
Wei, J.; and Zou, K. 2019.
\newblock {EDA}: Easy Data Augmentation Techniques for Boosting Performance on
  Text Classification Tasks.
\newblock In \emph{Proceedings of the 2019 Conference on Empirical Methods in
  Natural Language Processing and the 9th International Joint Conference on
  Natural Language Processing (EMNLP-IJCNLP)}, 6382--6388. Hong Kong, China:
  Association for Computational Linguistics.

\bibitem[{Xie et~al.(2019)Xie, Dai, Hovy, Luong, and
  Le}]{9-xie2019unsupervised}
Xie, Q.; Dai, Z.; Hovy, E.; Luong, M.-T.; and Le, Q.~V. 2019.
\newblock Unsupervised data augmentation for consistency training.
\newblock \emph{arXiv preprint arXiv:1904.12848}.

\bibitem[{Xie et~al.(2017)Xie, Wang, Li, L{\'e}vy, Nie, Jurafsky, and
  Ng}]{16-xie2017data}
Xie, Z.; Wang, S.~I.; Li, J.; L{\'e}vy, D.; Nie, A.; Jurafsky, D.; and Ng,
  A.~Y. 2017.
\newblock Data noising as smoothing in neural network language models.
\newblock \emph{arXiv preprint arXiv:1703.02573}.

\bibitem[{Yu et~al.(2018)Yu, Dohan, Luong, Zhao, Chen, Norouzi, and
  Le}]{back_translation1}
Yu, A.~W.; Dohan, D.; Luong, M.-T.; Zhao, R.; Chen, K.; Norouzi, M.; and Le,
  Q.~V. 2018.
\newblock Qanet: Combining local convolution with global self-attention for
  reading comprehension.
\newblock \emph{arXiv preprint arXiv:1804.09541}.

\bibitem[{Yu and Jiang(2016)}]{yu2016learning}
Yu, J.; and Jiang, J. 2016.
\newblock Learning sentence embeddings with auxiliary tasks for cross-domain
  sentiment classification.
\newblock Association for Computational Linguistics.

\bibitem[{Zhang et~al.(2018)Zhang, Cisse, Dauphin, and
  Lopez-Paz}]{zhang2018mixup}
Zhang, H.; Cisse, M.; Dauphin, Y.~N.; and Lopez-Paz, D. 2018.
\newblock mixup: Beyond Empirical Risk Minimization.
\newblock In \emph{International Conference on Learning Representations}.

\bibitem[{Zhang, Zhao, and LeCun(2015)}]{data-zhang-cnn}
Zhang, X.; Zhao, J.; and LeCun, Y. 2015.
\newblock Character-level convolutional networks for text classification.
\newblock \emph{Advances in neural information processing systems}, 28.

\end{thebibliography}



\bigskip

\end{document}